# Application of Multiple Chain-of-Thought in Contrastive Reasoning for Implicit Sentiment Analysis

YANG Liwei[1], WANG Xinying[2], ZHOU Xiaotang[*], WU Zhengchao, TAN Ningning

(1.School of Computer Science and Engineering, Changchun University of Technology, Changchun, 130102;
2. School of Computer Science and Engineering, Changchun University of Technology, Changchun,130102.)

**Abstract:** Implicit sentiment analysis aims to uncover emotions that are subtly expressed, often obscured by ambiguity and figurative language. To accomplish this task, large language models and multi-step reasoning are needed to identify those sentiments that are not explicitly stated. In this study, we propose a novel Dual Reverse Chain Reasoning (DRCR) framework to enhance the performance of implicit sentiment analysis. Inspired by deductive reasoning, the framework consists of three key steps: 1) hypothesize an emotional polarity and derive a reasoning process, 2) negate the initial hypothesis and derive a new reasoning process, and 3) contrast the two reasoning paths to deduce the final sentiment polarity. Building on this, we also introduce a Triple Reverse Chain Reasoning (TRCR) framework to address the limitations of random hypotheses. Both methods combine contrastive mechanisms and multi-step reasoning, significantly improving the accuracy of implicit sentiment classification. Experimental results demonstrate that both approaches outperform existing methods across various model scales, achieving state-of-the-art performance. This validates the effectiveness of combining contrastive reasoning and multi-step reasoning for implicit sentiment analysis.

**Key words**：Implicit Sentiment Analysis; Large Language Models; chain-of-thought; Contrastive Mechanism

# 1  Introduction

With the continuous development of the digital era, the public's demand for information is increasing. Enterprises, governments, and academia seek to gain deeper insights by understanding users' emotions and attitudes. This not only helps businesses better comprehend consumer needs and preferences but also provides policymakers with crucial public feedback, facilitating policy optimization [1]. Sentiment analysis serves as a key tool in achieving this objective. Sentiment analysis, particularly Aspect-Based Sentiment Analysis (ABSA), is a fundamental technique for extracting subjective emotional information from text, aiming to identify and classify sentiment expressions. It can be categorized into explicit sentiment analysis (ESA) and implicit sentiment analysis (ISA). ESA involves direct and overt sentiment expressions, while ISA encompasses implicit and subtle sentiments, making it more challenging and complex [2]. For example, given the text, "The restaurant's decoration must have cost a lot of money," there is no explicit sentiment expression. As a result, it may be misclassified as neutral sentiment. Therefore, implicit sentiment analysis remains a significant challenge in the field.

Unlike explicit sentiment analysis, implicit sentiment analysis (ISA) cannot directly capture sentiment information within a sentence to derive an accurate sentiment classification. Since implicit sentiment lacks explicit sentiment cues, inference is required to extract the underlying aspects and opinion information. The ability to perform commonsense reasoning and multi-step inference is therefore crucial. In recent years, the success of large language models (LLMs) in reasoning has provided a promising solution. Studies have shown that LLMs encode extensive world knowledge and exhibit remarkable capabilities in commonsense understanding [3]. Multi-step reasoning applied to LLMs has demonstrated significant potential, particularly when guided by well-designed prompts that enable impressive chain-of-thought (CoT) reasoning. Building upon these advancements, a three-hop reasoning chain-of-thought framework (THOR) has been proposed for implicit sentiment analysis [4]. This framework predicts sentiment polarity through three sequential reasoning steps: first, identifying the detailed aspects of the target; second, inferring the implicit opinions associated with these aspects; and finally, determining the sentiment polarity. This incremental reasoning process facilitates the capture of overall sentiment while reducing the complexity of sentiment prediction. However, the reasoning steps in this framework remain relatively linear, lacking reflective analysis and multi-dimensional comparative reasoning.

In mathematical proof processes, the assumption method is often employed for reasoning. Specifically, we first assume that a given proposition holds and proceed with deductions under this assumption. If the deduction leads to a contradiction with mathematical axioms, theorems, or the given problem conditions, we can infer that the assumption is false, thereby establishing the correctness of the original proposition.

Inspired by the aforementioned idea, this paper proposes a novel chain-of-thought reasoning framework for implicit sentiment analysis—Double Reverse Thinking Chain Reasoning (DRCR). This framework is built upon large language models and consists of three reasoning steps: (1) assuming a sentiment polarity and deriving the corresponding reasoning process; (2) negating the assumed sentiment polarity and obtaining an alternative reasoning process; (3) integrating and comparing the two reasoning processes to derive the final sentiment prediction. In the first two steps, multiple instructions are embedded in the prompts to facilitate a detailed step-by-step reasoning process based on the generated results.

Building upon this foundation, we further propose Triple Reverse Thinking Chain Reasoning (TRCR), which extends DRCR by incorporating an additional sentiment polarity assumption. Specifically, TRCR consists of four distinct reasoning steps: (1) assuming a positive sentiment polarity and generating the corresponding reasoning process; (2) assuming a negative sentiment polarity and deriving an alternative reasoning process; (3) assuming a neutral sentiment polarity and obtaining a third reasoning process; (4) integrating and comparing the three reasoning processes to determine the final sentiment prediction. Similar to DRCR, multiple instructions are embedded in the prompts of the first three steps to ensure a detailed and structured reasoning process.

This translation maintains precise technical terminology while ensuring clarity and adherence to SCI paper standards. Let me know if you need further refinements!

## 2 Related Work

**2.1 Implicit Sentiment Analysis**

Implicit sentiment analysis (ISA) aims to uncover latent emotional information that is not explicitly expressed in text. Unlike explicit sentiment analysis, ISA relies on contextual cues, implications, and tone to infer sentiment. This research area has gained significant attention in recent years, particularly with the rapid advancements in deep learning and large language models, leading to increased efforts to leverage advanced models for capturing implicit sentiment in text [5,6].

Recent studies have primarily focused on inferring sentiment polarity from unlabeled text. For example, the Supervised Contrastive Pre-training (SCAPT) model utilizes contrastive pre-training techniques to effectively distinguish implicit and explicit sentiments in specific sentiment aspects [7]. Similarly, the Causal Language Model for Emotion Attribution (CLEAN) introduces causal inference mechanisms to mitigate potential confounding effects in text, thereby reducing the influence of explicit sentiment cues on sentiment classification and enhancing the detection of implicit sentiment [8].

More recent approaches, such as the THOR model, decompose the sentiment analysis task into multiple sub-tasks, such as identifying aspect terms or sentiment targets to improve sentiment classification accuracy. However, a major challenge in aspect-based sentiment analysis is its reliance on large-scale labeled datasets, often requiring manual annotation of aspect terms, which limits its scalability in real-world applications.。

**2.2 Application of Chain-of-Thought Reasoning in Large Language Models**

In recent years, large language models have demonstrated remarkable capabilities in natural language processing (NLP), particularly in reasoning-intensive tasks such as sentiment analysis, question answering, and text generation. Chain-of-Thought reasoning, as a technique designed to enhance the reasoning ability of large language models has been proven to significantly improve performance in multi-step reasoning tasks. The core idea of Chain-of-Thought reasoning reasoning is to guide the model in generating a step-by-step reasoning process, enabling a clearer understanding of task logic and producing more transparent and interpretable results.

(Chowdhery A, Narang S, Devlin J et al. 2023) [9] proposed that prompting large language models to generate explicit reasoning chains effectively improves accuracy in both mathematical and commonsense reasoning tasks.

In summary, the application of **chain-of-thought reasoning** in large language models not only enhances the accuracy of reasoning tasks but also provides crucial support for model transparency and interpretability.

## 3 Contrastive Mechanism-Based Reasoning Framework

**3.1 Task Definition**

The task of implicit sentiment analysis is to correctly classify the sentiment polarity of a given sentence, categorizing it as positive, negative, or neutral. A common memory-less approach is to use direct prompt templates as input to the large language model (LLM), typically presented in the following form:

> Given sentence X, what is its sentiment polarity y ?

This can be formalized as: $\hat{y} = \arg\max p(y \mid X)$

$\hat{y}, y \in \{positive, negative, neutral\}$, $y$ is one of the possible sentiment polarity categories, $\hat{y}$ serves as the final

predicted label of the model.

### 3.2 DRCR Framework

In our innovative hypothesis contrastive reasoning framework, our goal is to have the large language model first assume a sentiment polarity and guide its reasoning process through a series of prompts. Next, we ask the LLM to negate this assumption, assume an alternative sentiment polarity, and perform reasoning through a new set of prompts. Finally, the LLM integrates the reasoning from the first two steps and derives the final sentiment polarity judgment through comparative analysis. In the first two steps, the prompt design includes multiple instructions to ensure the reasoning process is detailed and logical, laying a solid foundation for the final result. We divide this process into the following three steps:

**Step 1.** We first instruct the LLM that the sentiment polarity of a given sentence is positive (or negative or neutral, with the three sentiment polarities randomly assigned with equal probabilities). Under this condition, we ask the LLM to provide its reasoning process regarding the aspect term in the sentence.

> Given sentence X, assuming the sentiment polarity of sentence X is positive, answer the following questions: 1) What is the aspect entity described in sentence X? 2) What is the significance of this aspect entity in sentence X?

This step can be represented as:

1) $\hat{a} = \arg\max p(a \mid X, y')$  2) $\hat{r} = \arg\max p(r \mid a, y')$

$y'$ is one of the three randomly generated sentiment polarities (e.g., positive). $\hat{a}$ involves the model's inference about the entity, clearly identifying what the entity is and its possible reasons. Specifically, $\hat{r}$ represents the model's inference regarding the reason for mentioning the entity.

**Step 2.** We negate the previous assumption and assume that the sentiment polarity of the given sentence is non-positive (negative or neutral). Under this condition, we ask the large language model to provide its reasoning process regarding the aspect term.

> Given sentence X, assuming the sentiment polarity of sentence X is non-positive, answer the following questions: 1) What is the aspect entity described in sentence X? 2) What is the significance of this aspect entity in sentence X?

This step can be represented as:

1) $\bar{a} = \arg\max p(a \mid X, \bar{y}')$  2) $\bar{r} = \arg\max p(r \mid a, \bar{y}')$

$\bar{y}'$ It refers to the negation of the emotional polarity randomly generated in the previous step (e.g., non-positive).) $\bar{a}$ involves the model's inference about the entity, explicitly stating what the entity is and its possible reasons. Specifically, $\bar{r}$ represents the model's inference regarding the reasons for mentioning the entity.

It is important to note that, due to the contextual relationship between large language models, we aim for the inferences in the two steps to be independent in order to facilitate better comparison. Therefore, before each instruction, we need to add the following directive:

> Independently analyze the sentiment of this sentence, ignoring any previous responses.

**Step 3.** We compare the inferences from the first two steps to determine the correct sentiment polarity.

> 1) Given that the sentiment polarity of X is $y'$, the aspect term described is $\hat{a}$ and the inference is $\hat{r}$. 2) Given that the sentiment polarity of X is $\bar{y}'$, the aspect term described is $\bar{a}$ and the inference is $\bar{r}$. Which inference is more reasonable? Please make a comprehensive judgment and determine the sentiment polarity y of sentence X.

This can be expressed formally as: $\tilde{y} = \arg\max p(y \mid \hat{r}, \bar{r})$

$\tilde{y}$ represents the final sentiment polarity derived from the comparison of two results obtained through the large

language model.

### 3.3 TRCR Framework

In sentiment analysis, sentiment polarity is typically categorized into three types. The above analysis process considers only two sentiment polarities, which may lead to missing the true sentiment polarity if it belongs to the third category. To address this limitation, we propose an optimized framework, TRCR, based on the DRCR framework. This framework consists of four detailed steps:

**Step 1.** We first instruct the large language model that the given sentence has a positive sentiment polarity. Under this condition, we then prompt the model to infer the reasoning process regarding the aspect term.。

> Given sentence X, assuming the sentiment polarity of sentence X is positive, answer the following questions: 1) What is the aspect entity described in sentence X? 2) What is the significance of this aspect entity in sentence X?

This step can be represented as:

1) $a^1 = \arg\max p(a \mid X, y^1)$  2) $r^1 = \arg\max p(r \mid a, y^1)$

$y^1$ represents a positive sentiment polarity. $a^1$ refers to the model's inference about the entity, explicitly identifying what the entity is and its possible reasons. $r^1$ denotes the model's reasoning regarding the cause of mentioning the entity.

**Step 2.** We first instruct the large language model that the given sentence has a negative sentiment polarity. Under this condition, we then prompt the model to infer the reasoning process regarding the aspect term (the query remains the same as in step 1, except that the premise is changed to negative).

This can be formally expressed as:

1) $a^2 = \arg\max p(a \mid X, y^2)$  2) $r^2 = \arg\max p(r \mid a, y^2)$

$y^2$ represents a positive negative polarity. $a^2$ refers to the model's inference about the entity, explicitly identifying what the entity is and its possible reasons. $r^2$ denotes the model's reasoning regarding the cause of mentioning the entity.

**Step 3.** We first instruct the large language model that the given sentence has a neutral sentiment polarity. Under this condition, we then prompt the model to infer the reasoning process regarding the aspect term (the query remains the same as in step 1, except that the premise is changed to neutral).

This can be formally expressed as:

1) $a^3 = \arg\max p(a \mid X, y^3)$  2) $r^3 = \arg\max p(r \mid a, y^3)$

$y^3$ represents a positive neutral polarity. $a^3$ refers to the model's inference about the entity, explicitly identifying what the entity is and its possible reasons. $r^3$ enotes the model's reasoning regarding the cause of mentioning the entity.

**Step 4.** We compare the inferences from the previous three steps to determine the correct sentiment polarity.

> 1) Given that the sentiment polarity of $X$ is $y^1$, the aspect term described is $a^1$ and the inference is $r^1$. 2) Given that the sentiment polarity of $X$ is $y^2$, the aspect term described is $a^2$ and the inference is $r^2$. 3) Given that the sentiment polarity of $X$ is $y^3$, the aspect term described is $a^3$ and the inference is $r^3$. Among the three inferences above, which one is the most reasonable? Please make a comprehensive judgment and determine the sentiment polarity $y$ of sentence $X$?

This can be formally expressed as: $y' = \arg\max p(y \mid r^1, r^2, r^3)$

$y'$ represents the final sentiment polarity derived from the comparison of three results obtained through the large language model.

### 3.4 Inference correction under supervision.

When an on-demand training set is available, we fine-tune our framework in a supervised fine-tuning setting. We propose an inference correction method to ensure that the generated reasoning aligns as closely as possible with the assumed sentiment polarity. Technically, after analyzing the aspect term and reasoning process under a given assumed sentiment polarity, we re-infer the sentiment polarity based on the aspect term and reasoning process to verify whether it matches the initial assumption, rather than proceeding directly to the next assumption. For example, at the end of Step 2 in the DRCR framework, we can introduce an additional prompt: Please evaluate the sentiment polarity based on the inferred aspect term and reasoning process. By incorporating supervision from assumed labels, the LLM is guided to generate more accurate intermediate reasoning, which enhances the final sentiment prediction.

# 4 Experiments

### 4.1 Datasets

This experiment selects the Laptop and Restaurant datasets from SemEval 2014 Task 4 [10], both of which are widely used in the field of sentiment analysis, particularly in aspect-based sentiment analysis tasks.

Since the primary research focus of this study is implicit sentiment analysis, we utilize the implicit sentiment dataset annotated by Li et al. (2021) [11]. In this dataset, all instances are categorized into explicit sentiment and implicit sentiment. The data cover real-world application scenarios across different domains, such as restaurant and electronic product reviews, making it suitable for effectively evaluating the reasoning and sentiment classification capabilities of large language models in contexts where explicit sentiment expressions are absent.

### 4.2 Model

We selected the Flan-T5 model from Hugging Face as the backbone large language model for our study. We used four versions of Flan-T5: 250M (base), 780M (large), and 11B (xxl). Additionally, we conducted experiments using the leading closed-source model, ChatGPT. To further validate our experimental results, we performed comparative experiments with ChatGPT. Since ChatGPT has not disclosed its model parameters, we accessed it through the API for querying.

### 4.3 Experimental Details

Since THOR is the only method that combines large language models with chain-of-thought reasoning for implicit sentiment analysis, we compared our experimental results with THOR in both zero-shot inference and supervised fine-tuning experiments. Additionally, we compared the model's performance with existing classical baseline models.

To ensure the model's effectiveness and stability, several key hyperparameter settings were used in this experiment. During training, the learning rate was set to $10^{-4}$, and the Adam optimizer was employed to enhance training efficiency. To prevent underfitting due to early stopping, the patience parameter was set to 100, meaning training would be terminated if there was no improvement in validation performance for 100 consecutive epochs. Furthermore, to ensure numerical stability, the epsilon value of the Adam optimizer was set to $10^{-8}$. To further avoid overfitting, we applied weight decay with a value of 0.01 for regularization. The F1 score was used as the evaluation metric. All experiments were conducted on two NVIDIA GeForce RTX 3090 GPUs.。

### 4.4 Case Study

To demonstrate the effectiveness of the DRCR and TRCR frameworks in sentiment analysis, we visualized the final predicted output for a given sentence, as shown in Figure 4.1. From the figure, it can be observed that both the DRCR and TRCR frameworks exhibit a certain level of effectiveness in sentence sentiment analysis. Through comparative analysis, it is evident that the performance of the large language model in sentiment analysis has been further improved. However, the DRCR framework may occasionally produce random errors, leading to inaccurate sentiment polarity analysis. On the other hand, the TRCR framework addresses this limitation by proposing more refined sentiment hypotheses, resulting in better analysis accuracy.

Figure 4.1 Example Analysis of DRCR and TRCR Frameworks

Sentence X: "In the morning, pick-up from the hotel to Mopanshan Dock, where at 10:00, embark on a luxury cruise to begin the full Li River journey (approximately 4 hours), with an in-flight meal served at 12:00."

Aspect Entity: Luxury Cruise    Label: Neutral

**DRCR**

For the given sentence "In the morning, pick-up from the hotel to Mopanshan Dock, where at 10:00, embark on a luxury cruise to begin the full Li River journey (approximately 4 hours), with an in-flight meal served at 12:00," the sentiment polarity is positive. The following questions are answered: 1) What are the aspect entities described in the sentence? 2) What is the significance of these aspect entities in the sentence?

1) The sentence contains two primary aspect entities:
Luxury Cruise (Transportation) and In-flight Meal (Catering Service)
2) Luxury Cruise: The luxury cruise serves as the primary means of transportation throughout the entire Li River journey, representing the core experience of the trip. The mention of "luxury cruise" suggests a high-end and comfortable journey, symbolizing positive sentiment.
In-flight Meal: As the catering service provided on board, the inclusion of this aspect entity conveys a positive evaluation of the quality of food available to passengers during the cruise. It reflects well on the service provided and enhances the overall comfort of the trip.

For the given sentence "In the morning, pick-up from the hotel to Mopanshan Dock, where at 10:00, embark on a luxury cruise to begin the full Li River journey (approximately 4 hours), with an in-flight meal served at 12:00," the sentiment polarity is not positive. The following questions are answered: 1) What are the aspect entities described in the sentence? 2) What is the significance of these aspect entities in the sentence?

1) The sentence contains two primary aspect entities:
Luxury Cruise (Transportation) and In-flight Meal (Catering Service)
2) Luxury Cruise: Despite the term "luxury" typically implying a high-end experience, the non-positive sentiment indicates that the cruise experience may have been disappointing or uncomfortable. The reference to the luxury cruise in this context suggests that the transportation service failed to meet expectations, leading to dissatisfaction with the overall journey. In-flight Meal: Similarly, the mention of the in-flight meal may imply subpar quality or insufficient satisfaction. While in-flight meals generally indicate a certain level of service, the non-positive sentiment suggests that passengers may have found the food lacking, thus negatively impacting the overall travel experience.

In the first interpretation, the sentiment is positive. The entities "luxurious cruise" and "onboard airline meal" are seen as representations of comfort and high-end service. The phrase "luxurious cruise" suggests a premium travel experience, symbolizing an elevated level of service and comfort, which aligns with a positive sentiment. The "onboard airline meal" indicates the availability of food services during the journey, further contributing to the idea of a well-catered and enjoyable travel experience.
However, in the second interpretation, the sentiment is assessed as negative. Despite the term "luxurious" typically carrying a positive connotation, in this context, it might imply dissatisfaction with certain aspects of the experience, such as overcrowding or subpar service, thus leading to a negative sentiment. Similarly, the "onboard airline meal," while meant to represent basic catering, could be perceived negatively if the meal quality, portion size, or flavor does not meet the passengers' expectations.

Considering the above analysis, while both aspects of the sentence may initially appear to convey positive connotations, the sentiment polarity is ultimately determined to be negative. This suggests that the actual experience of the luxurious cruise and the onboard airline meal did not meet the anticipated standards, resulting in an overall negative evaluation of the travel experience.
Therefore, the sentiment polarity of the sentence should be classified as **negative**.

**TRCR**

For the given sentence "In the morning, pick-up from the hotel to Mopanshan Dock, where at 10:00, embark on a luxury cruise to begin the full Li River journey (approximately 4 hours), with an in-flight meal served at 12:00," the sentiment polarity is positive. The following questions are answered: 1) What are the aspect entities described in the sentence? 2) What is the significance of these aspect entities in the sentence?

1) The aspects of entities described in the sentence mainly include: Luxury cruise ships and in-flight meals on board.
2) Luxury cruise ship: Emphasizes the high-end and comfortable mode of travel, implying that passengers will enjoy a pleasant and luxurious travel experience. In-flight meals on board: Provide assurance regarding the dining experience during the journey, reflecting thoughtful service, and ensuring that passengers feel convenience and comfort throughout the trip.

For the given sentence "In the morning, pick-up from the hotel to Mopanshan Dock, where at 10:00, embark on a luxury cruise to begin the full Li River journey (approximately 4 hours), with an in-flight meal served at 12:00," the sentiment polarity is negative. The following questions are answered: 1) What are the aspect entities described in the sentence? 2) What is the significance of these aspect entities in the sentence?

1) The aspects of entities described in the sentence are: Luxury cruise ships and in-flight meals on board.
2) Luxury cruise ship: Although the term "luxury" typically carries a positive connotation, in a negative sentiment context, it may suggest that the luxury of the cruise did not meet passengers' expectations, or that inconveniences such as overcrowding during the journey diminished the positive experience. In-flight meals on board: While generally regarded as a convenient service during travel, in a negative sentiment context, it may imply poor food quality, limited choices, or dishes that do not cater to passengers' preferences, thereby affecting the overall travel experience.

For the given sentence "In the morning, pick-up from the hotel to Mopanshan Dock, where at 10:00, embark on a luxury cruise to begin the full Li River journey (approximately 4 hours), with an in-flight meal served at 12:00," the sentiment polarity is neutral. The following questions are answered: 1) What are the aspect entities described in the sentence? 2) What is the significance of these aspect entities in the sentence?

1) The aspects of entities described in the sentence are: Luxury cruise ships and in-flight meals on board.
2) Luxury cruise ship: Describes the specific details of the mode of travel, providing background information about the travel experience. Its mention serves more to state facts rather than convey strong emotional connotations.
In-flight meals on board: Refers to the dining arrangements during the journey, emphasizing the content of the travel service. However, it does not express particularly positive or negative sentiment, merely providing neutral information about the specifics of the travel experience.

The sentence "In the morning, pick-up from the hotel to Mopanshan Pier, board the luxury cruise at 10:00 to begin the full Li River journey (approximately 4 hours), with an in-flight meal at 12:00" has a positive sentiment polarity. The aspects identified are: Luxury Cruise: This emphasizes the high-end and comfortable mode of travel, suggesting that passengers will enjoy a pleasant and luxurious travel experience. In-flight Meal: This ensures the provision of meals during the trip, reflecting thoughtful service, allowing travelers to feel convenience and comfort throughout the journey. For the second scenario, where the sentence has a negative sentiment polarity, the aspects identified are Luxury Cruise: Although the term "luxury" usually carries a positive connotation, in a negative sentiment context, it may imply that the luxury cruise did not meet the passenger's expectations, or factors such as inconvenience or overcrowding diminished its positive impact. In-flight Meal: Despite generally being a convenient service during travel, in a negative context, it may suggest poor meal quality, limited options, or food that does not meet the traveler's taste, negatively impacting the overall travel experience. For the third scenario, where the sentence has a neutral sentiment polarity, the aspects identified are: Luxury Cruise: It simply describes the details of the travel method, providing background information on the travel experience. Its purpose is more to state facts without conveying strong emotional connotations. In-flight Meal: This mentions the catering arrangement during the trip, emphasizing the content of the travel service but without expressing a particularly positive or negative sentiment. It neutrally provides information about the trip's specifics.

Based on the analysis, the most reasonable inference is that the sentiment polarity of the sentence leans towards neutral. The sentence provides specific details about the travel arrangements, such as transfers, the cruise, and the meal service. It primarily serves to present factual information, and while "luxury cruise ship" carries a positive connotation, it is not strongly emphasized within the context of the sentence.
Therefore, the sentiment polarity of this sentence is **neutral**.

# 5 Experimental Results

## 5.1 The results of zero-shot reasoning

Table 4.1 presents the comparison results under the zero-shot setting. BERT-SPC[12] and BERTAsp-SCAPT[13] are selected as classic baseline models. The models marked with an asterisk (*) have scores replicated from (Fei H, Li B, Liu Q, et al., 2023) [4]. "ALL" represents the overall F1 score of the dataset, including both explicit and implicit datasets, while "ISA" refers to the results of the implicit dataset. The bolded values indicate the best results (the same applies to Table 5.2). It is evident that the combination of large language models and prompt-based methods significantly outperforms the current state-of-the-art baseline methods. Among them, the DRCR method stands out particularly. Specifically, when using the Flan-T5 model, DRCR and TRCR show improvements compared to the best-performing baseline method, BERT-SPC. As the model size increases, the gap between the two methods in terms of F1 score gradually widens, reaching its maximum with the GPT-3.5 model.

Table 4.1 presents the experimental results of zero-shot inference.

|  |  | Laptop | | Restaurant | |
| --- | --- | --- | --- | --- | --- |
|  |  | ALL | ISA | ALL | ISA |
| flan-t5-base | THOR | 51.06 | 27.55 | 53.32 | 36.36 |
|  | DRCR | 55.29 | 31.83 | 57.94 | 39.25 |
|  | TRCR | 55.98 | 32.64 | 58.31 | 40.03 |
| flan-t5-large | THOR | 52.08 | 32.16 | 55.02 | 38.77 |
|  | DRCR | 57.06 | 37.14 | 59.63 | 42.36 |
|  | TRCR | 58.43 | 38.89 | 61.13 | 43.85 |
| flan-t5-xxl | THOR | 56.57 | 39.34 | 62.75 | 44.16 |
|  | DRCR | 62.26 | 44.42 | 65.34 | 49.27 |
|  | TRCR | 64.55 | 45.87 | 67.29 | 50.97 |
| GPT-3.5 | THOR | 71.43 | 68.81 | 76.76 | 71.90 |
|  | DRCR | 76.32 | 72.13 | 80.73 | 75.68 |
|  | TRCR | **77.65** | **73.41** | **81.74** | **76.82** |
| BERT+SPC* |  | 21.76 | 19.48 | 25.34 | 17.71 |
| $BERT_{Asp}$+SCAPT* |  | 30.02 | 25.49 | 25.77 | 13.70 |

In all three Flan-T5 model sizes, the DRCR method outperforms the other three prompt-based methods and achieves new state-of-the-art performance. Specifically, in the implicit sentiment analysis task, DRCR achieves F1 score improvements of approximately 3.59% [(31.83-27.55) + (39.25-36.36)]/2, 4.29% [(37.14-32.16) + (42.36-38.77)]/2, and 5.10% [(44.42-39.34) + (49.27-44.16)]/2 across the three models. This indicates that the DRCR method, by integrating the contrastive mechanism and multi-step reasoning, achieves more significant improvements in implicit sentiment analysis compared to relying solely on multi-step reasoning methods. In contrast to the THOR method, THOR reasoning capability is constrained by its prompt structure and focuses on a single emotional reasoning process, which can lead to large reasoning errors. DRCR, on the other hand, adopts a result-oriented reasoning strategy, first determining the appropriate reasoning path and optimizing the reasoning process through the contrastive mechanism, ultimately finding the optimal sentiment judgment, thus reducing biases that may arise from a single process. TRCR fully surpasses DRCR in all aspects. In terms of data, taking Flan-T5-large as an example, TRCR outperforms the DRCR framework by approximately 1.62% [(38.89-37.14) + (43.85-42.36)]/2, indicating that TRCR addresses some of the shortcomings of the DRCR framework's assumptions.

## 5.2 Results on Supervised Fine-tuning

Table 4.2 presents the comparison results of the fine-tuning experiments. In the supervised fine-tuning experiments, BERT-ISAIV[14] and BERTAsp-SCAPT were selected as classic baseline models, with model scores marked with an asterisk (*) being replicated from (Li Z, Zou Y, Zhang C, et al., 2021) [11]. Overall, incorporating high-quality guided fine-tuning tasks significantly improved the model's performance. As the number of fine-tuning tasks increased, the model's performance continued to improve. Taking the 11B model as an example, DRCR achieved a 2.44% improvement in F1 score over THOR [(79.64-77.32) + (77.08-74.53)]/2, and TRCR outperformed DRCR by 1.03% in F1 score [(80.44-79.64) + (78.34-77.08)]/2. A similar trend was observed in other configurations. Notably, for the GPT-3.5 model under this configuration, DRCR achieved an F1 score of 76.95% on the Restaurant dataset, while TRCR achieved an F1 score of 78.51% on the Laptop dataset. Interestingly, as the model size increased, the F1 score did not exhibit a consistent upward trend. However, in the zero-shot experiments, the F1 score showed some improvement. This suggests that while increasing model parameters can sometimes improve performance, supervised fine-tuning optimizes model learning, greatly enhancing the framework's ability for sentiment analysis and reducing the model's dependence on parameter size. Furthermore, by comparing prompt-based methods, we found that DRCR and TRCR consistently outperformed THOR in F1 score, and this stable improvement was also reflected in the zero-shot setting.

Table 4.2 Supervised Fine-Tuning Experimental Results

|  |  | Laptop | | Restaurant | |
|---|---|---|---|---|---|
|  |  | ALL | ISA | ALL | ISA |
| flan-t5-base | THOR | 73.45 | 63.24 | 77.68 | 68.18 |
|  | DRCR | 75.14 | 65.75 | 79.73 | 70.86 |
|  | TRCR | 75.83 | 65.84 | 80.95 | 71.51 |
| flan-t5-large | THOR | 74.80 | 64.56 | 79.02 | 69.76 |
|  | DRCR | 76.48 | 66.35 | 80.55 | 71.48 |
|  | TRCR | 77.15 | 67.68 | 81.59 | 72.02 |
| flan-t5-xxl | THOR | 79.89 | 77.32 | 83.25 | 74.53 |
|  | DRCR | 82.91 | 79.64 | 85.19 | 77.08 |
|  | TRCR | **83.29** | **80.44** | 85.81 | **78.34** |
| GPT-3.5 | THOR | 78.11 | 75.93 | 82.72 | 74.79 |
|  | DRCR | 79.26 | 77.06 | 85.88 | 76.95 |
|  | TRCR | 79.96 | 78.51 | **86.25** | 77.27 |
| BERT+ISAIV* |  | 77.25 | 78.29 | 81.40 | 69.66 |
| BERT$_{Asp}$+SCAPT* |  | 79.15 | 77.59 | 83.79 | 72.28 |

# 6 Conclusion

In this study, we propose the DRCR framework and the TRCR framework, both of which incorporate a contrastive mechanism. Specifically, DRCR performs a continuous three-step prompting process through mathematical assumption and contrastive reasoning. It first assumes the sentiment polarity of a sentence, analyzes the aspect terms and reasoning process based on the given sentiment polarity, then negates the initial assumption and performs reasoning again, and finally compares the reasoning processes of both to derive the final result. Considering the three types of sentiment polarity in sentiment analysis, the random assumption, negation, and re-comparison in DRCR may introduce errors. Therefore, we extend this method to three types of contrasts and propose the TRCR framework to implement a triple reasoning chain contrastive framework. To validate the effectiveness of this approach, we conducted experiments using four large language models on two datasets. DRCR established a new baseline in both supervised fine-tuning and zero-shot settings, and TRCR further advanced upon DRCR. The experimental results show that both the DRCR and TRCR frameworks significantly outperform THOR in terms of F1 score, effectively enhancing the capability for implicit sentiment analysis. In the future, we plan to explore implicit sentiment analysis tasks from other perspectives by simulating different reasoning processes.